\pdfoutput=1

\documentclass[11pt]{article}

\usepackage{emnlp2021}

\usepackage{times}
\usepackage{latexsym}

\usepackage[T1]{fontenc}

\usepackage[utf8]{inputenc}

\usepackage{microtype}
\usepackage{color,soul}

\usepackage{graphicx}
\usepackage{booktabs}
\usepackage{subcaption}
\usepackage{url}
\usepackage{covington}

\renewcommand{\vec}[1]{\mathbf{#1}}

\title{Translation Error Detection as Rationale Extraction}

\author{\\
	\textbf{Marina Fomicheva\textsuperscript{1} \qquad Lucia Specia\textsuperscript{1,2}  \qquad Nikolaos Aletras\textsuperscript{1}}\\
	\textsuperscript{1}Department of Computer Science, University of Sheffield \\
	\textsuperscript{2}Department of Computing, Imperial College London \\
	                   United Kingdom \\
	\texttt{\{m.fomicheva,l.specia,n.aletras\}@sheffield.ac.uk}\\
}

\begin{document}
\maketitle
\begin{abstract}
Recent Quality Estimation (QE) models based on multilingual pre-trained representations have achieved very competitive results when predicting the overall quality of translated sentences. Predicting translation errors, i.e. detecting specifically which words are incorrect, is a more challenging task, especially with limited amounts of training data. We hypothesize that, not unlike humans, successful QE models rely on translation errors to predict overall sentence quality. 
By exploring a set of feature attribution methods that assign relevance scores to the inputs to explain model predictions, we study the behaviour of state-of-the-art sentence-level QE models and show that explanations (i.e. rationales) extracted from these models can indeed be used to detect translation errors. We therefore (i) introduce a novel semi-supervised method for word-level QE and (ii) propose to use the QE task as a new benchmark for evaluating the plausibility of feature attribution, i.e. how interpretable model explanations are to humans.
\end{abstract}


\section{Introduction}

Quality Estimation (QE) is the task of predicting Machine Translation (MT) quality at inference time, when no gold standard human translation is available as reference \citep{qe_mt_blatz, qe_mt_specia}. QE can be framed as a word-level or a sentence-level task. Both tasks have numerous practical applications, such as deciding whether a given MT output can be published without editing, highlighting potential critical errors, etc. Current QE approaches proceed by fine-tuning powerful representations from pre-trained multilingual sentence encoders such as BERT~\cite{devlin2018bert} or XLM-R~\cite{conneau2019unsupervised}. In the recent Shared Task on QE at WMT2020 \cite{specia-etal-2020-findings-wmt} these approaches have achieved very high performance at predicting sentence-level translation quality (up to 0.9 Pearson correlation with human judgements for some language pairs). However, as evidenced by the results from the mentioned shared task, the accuracy of word-level prediction still leaves room for improvement. This is partly due to the limited amount of training data. Word-level error annotation is especially time-consuming and expensive, as it requires work of bilingual experts. In this work we introduce a new semi-supervised approach to word-level QE that removes the need for training data at word level. To achieve this we propose to address QE as a rationale extraction task. 



Explainability is a broad area aimed at explaining predictions of machine learning models. Rationale extraction methods achieve this by selecting a portion of the input that justifies model output for a given data point. In translation, human perception of quality is guided by the presence of translation errors \cite{freitag2021experts}. We hypothesize that sentence-level QE models also rely on translation errors to make predictions. If that is the case, explanations for sentence-level predictions can be used to detect translation errors, thus removing the need for word-level training data. To extract model explanations, we use \textit{post hoc} rationale extraction methods \cite{sundararajan2017axiomatic} which try to explain the predictions of a given model (as opposed to modifying its architecture or introducing constraints during training), since one of our goals is to study to what extent existing QE models rely on the same information as humans.



At the same time, by treating word-level errors as explanations for sentence-level score we introduce a new benchmark for evaluating explainability methods.
Recent work has introduced various datasets for measuring the agreement between rationales extracted from NLP models and provided by human annotators \cite{deyoung2019eraser}. QE is different from these datasets in various important aspects. First, it is a regression task, as opposed to binary or multiclass text classification explored in previous work. Second, it is a multilingual task where the output score captures the relationship between source and target sentences. Finally, manual annotation of translation errors is a practical task with a long tradition in MT research and translation studies, and thus offers an interesting alternative to human explanations collected specifically for evaluating rationale extraction methods.


Our {\bf main contributions } are:
\begin{itemize}
    \item We introduce a novel semi-supervised method for word-level QE. We provide practical recipes on how feature attribution methods can be used to derive information on translation errors from sentence-level models. 
    \item We provide insights into the behaviour of QE models based on pre-trained Transformers by analysing attributions to different parts of input sequence (source vs. target sentence, correct words vs. errors) at different hidden layers.
    \item We propose to use the QE task as a new benchmark for evaluating the plausibility aspect of feature attribution, i.e. how interpretable model explanations are to humans.
\end{itemize}



\section{Background and Related Work}
\label{sec:background}

\paragraph{Quality Estimation}


%
Current SOTA in \textit{sentence-level} QE, which is typically framed as a regression task, explores multilingual representations from pre-trained Transformer models, notably XLM-Roberta. The input to sentence-level QE model with such an architecture is a concatenation of the source and translated sentence, separated by the [SEP] token. The sequence is encoded by the pre-trained Transformer model, and the [CLS] token is passed through a multilayer perceptron (MLP) layer to obtain a sentence-level score. During fine-tuning both the parameters of the pre-trained model and the parameters corresponding to the MLP layer are updated. \textit{Word-level} QE is typically addressed as a binary classification task, where for each word in the MT output the model needs to predict a binary label indicating whether the word is correct or corresponds to an error. For word-level similar architectures based on pre-trained Transformers are employed \cite{lee-2020-two}.
 
\begin{figure*}[t]
\centering
\begin{subfigure}[t]{0.45\textwidth}
    \includegraphics[width=.7\textwidth]{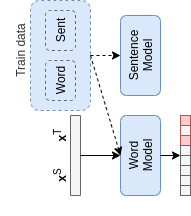}
\end{subfigure}%
\begin{subfigure}[t]{0.45\textwidth}
    \includegraphics[width=.7\textwidth]{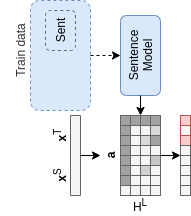}
\end{subfigure}
\caption{Fully supervised word-level QE (left) and semi-supervised word-level QE as rationale extraction (right). Dashed and solid lines represent training and test time, respectively.}
\label{fig:approach}
\end{figure*}

The vast majority of previous work has addressed word-level QE as a supervised task. As illustrated in Figure \ref{fig:approach} (left), some approaches use both sentence-level and word-level objectives in a multi-task setting, which results in superior performances \cite{kim-lee-na:2017:WMT,lee-2020-two}. Methods that do not require word-level training data either need access to the translation model \cite{rikters2017confidence,fomicheva2020unsupervised}, or still treat the problem as a supervised task but use synthetically generated data for supervision \cite{tuan2021quality}.




\paragraph{Rationale Extraction for NLP}

SOTA NLP models based on deep neural networks achieve high performances in a variety of tasks, often at the cost of interpretability \cite{lipton2016mythos}. Recent work on explainability aims to address this issue. Existing approaches focus on two different goals. On the one hand, they aim to produce justifications to model predictions that are \textit{plausible} to the users, in order to increase users' trust \cite{ribeiro2016should}. On the other hand, they aim to reveal the inner workings of the model and \textit{faithfully} explain model predictions, and thus can be useful to model developers \cite{jacovi2020towards}. 

Typically, explainability methods operate by selecting a portion of the input that justifies model prediction for a single data point. This can be done either by modifying model architecture for the purpose of rationale extraction, or by trying to explain the predictions of a given model. The first type of approaches (a.k.a. rationalization by construction) involves imposing restrictions on the generated rationales to satisfy certain constraints, e.g. compactness \cite{yu2019rethinking,chalkidis2021paragraph}. Note that such restrictions often result in lower performances and indeed are not guaranteed to explain the behaviour of an unconstrained model \cite{jain2020learning}. The second type of approaches (the so called \textit{post hoc} approaches) usually rely on feature attribution methods, which assign an attribution value to each input feature of a network \cite{sundararajan2017axiomatic,schulz2020restricting}. 
These methods do not allow for introducing useful biases during training, but focus on faithfully explaining model behaviour.

Feature attribution has a long tradition in the image recognition tasks \cite{simonyan2013deep}. Only recently, explanation techniques have been applied to some NLP tasks, most commonly text classification. QE is fundamentally different from text classification where clues are typically separate words or phrases \cite{zaidan2007using} which often can be considered independently of the rest of the text. This independence assumption does not hold for the task of evaluating translation quality where a word cannot be identified as a clue (e.g. translation error) without considering the surrounding context.

Furthermore, SOTA NLP models based on contextualized representations of input words make rationale extraction especially challenging, as the representation for a given word can encode not only the word identity but also its interactions with other words in the text. Recent work has revealed various interesting properties that characterize the flow of information through hidden layers in deep Transformer models \cite{voita2019bottom,de2020decisions,yun2021transformer}. We provide additional insights on this topic and discuss its relation to the aforementioned work.



\section{Translation Error Prediction as Rationale Extraction}
\label{sec:approach}

In this section we present our approach for semi-supervised word-level QE as rationale extraction.

\subsection{Approach}
As shown in Figure \ref{fig:approach} (right), instead of training a dedicated model for word-level prediction, we propose to derive word-level scores from a strong sentence-level model through rationale extraction. Given a trained model and the test data, rationale extraction methods detect the parts of the input that are relevant for model predictions on this data on a sample-by-sample basis. In this work, we argue that in the case of MT evaluation, words with the highest relevance scores should correspond to translation errors.

More formally, given the source sequence $\vec{x}^S=x_1^S,...,x_{|S|}^S$, the target sequence $\vec{x}^T=x_1^T,...,x_{|T|}^T$ and the QE model $M(\vec{x}^S,\vec{x}^T)=\hat{y}$ that predicts sentence quality, a feature attribution method produces a vector of attribution scores $\vec{a}=a_1,...,a_{|S+T|}$, which represent the contribution of each source and target word to the prediction $\hat{y}$.

Crucially, no word-level labels are required for training. For evaluation, the attribution scores are compared against binary gold labels $\vec{w}=w_1,...,w_{|T|}\in\{0,1\}$ indicating whether each given word in the target sequence is an error or is correct. 

The predictive models for QE explored in our experiments are built by fine-tuning multilingual representations from pre-trained Transformers. Transformer model starts from context-agnostic representations consisting of positional and token embeddings. These representations are passed through a set of hidden layers ($L$) where at each layer the representations $H_l={H_1^{(l)}, …, H_{|S+T|}^{(l)}}$ are iteratively updated via multi-head attention. This allows the hidden representation for each token to encode information on other words in the sentence. 

We note that attribution to the input tokens or to the embedding layer can hardly succeed in detecting translation errors, as those cannot be identified independently from the context given by the source and target sentence. In this work, we perform feature attribution to hidden states at different layers and analyse which layer results in attribution scores that best correspond to translation errors. 

\subsection{Explainability Methods}

Explainability methods can be divided into explanations by simplification, such as LIME \cite{ribeiro2016should}; gradient-based explanations \cite{sundararajan2017axiomatic}; and perturbation-based explanations \cite{schulz2020restricting}.

We select three popular methods for rationale extraction, which (i) do not require modifying the model architecture or re-training the model and (ii) allow attribution to hidden states. For comparison, we also use LIME which operates directly on the input text. We note that this set is not exhaustive or representative of SOTA rationale extraction methods. Our main goal is not conduct a comparative study but rather to test whether it is possible to address word-level QE as a rationale extraction task without any word-level supervision. 

\paragraph{LIME}
\cite{ribeiro2016should} is a simplification-based explanation technique, which fits a sparse linear model in the vicinity of each test instance, to approximate the decision boundary of the complex model.\footnote{We use the implementation available at \url{https://github.com/marcotcr/lime}} The data for fitting the linear model is produced by perturbing the given instance and computing model predictions. Linear model coefficients are then used as attribution scores for each input feature. For NLP tasks features correspond to input tokens and perturbation is achieved by randomly removing words from the sequence.

\paragraph{Information Bottleneck}
is a perturbation-based method originally proposed by \citet{schulz2020restricting} for the task of image recognition. The method applies the idea of information bottleneck \cite{tishby2015deep} for feature attribution. Specifically it injects noise into an intermediate layer representations. The amount of noise injected at the position corresponding to each input feature is optimized to minimize the loss of the main task while at the same time maximizing the overall amount of injected noise. 

\paragraph{Integrated Gradients}
\cite{sundararajan2017axiomatic} is a gradient-based method similar to the traditional salience and input$*$gradients approaches. The latter takes the signed partial derivatives of the output with respect to the input and multiply them by the input itself. Intuitively, this is analogous to inspecting the products of model coefficients and feature values in linear models \cite{sundararajan2017axiomatic}. Integrated gradients improves on that by defining a baseline input and computing the average gradient while the input varies along a linear path from baseline input to the actual input. The baseline is defined by the user depending on the task. For image recognition, black image is used as baseline. It is not clear what such baseline representation should be in the case of language tasks. Here, we select a zero baseline for simplicity. Better results can be achieved with a more informed choice of a baseline and we leave this to future work.\footnote{For both information bottleneck and integrated gradients method we adapt the implementation available at \url{https://github.com/nicola-decao/diffmask} for our QE scenario. Our version of the code will be made available upon acceptance.}

\paragraph{Attention}
Finally, we test attention as an attribution method. Self-attention mechanisms have been widely studied in the context of explainability \cite{jain2019attention,serrano2019attention,attention2019wiegreffe}. To compute a single attention scores for Transformer models with multi-head attention, we average the weights across the different attention heads. 

\begin{figure*}[ht]
\centering
\begin{subfigure}[t]{0.5\textwidth}
    \includegraphics[width=.9\textwidth]{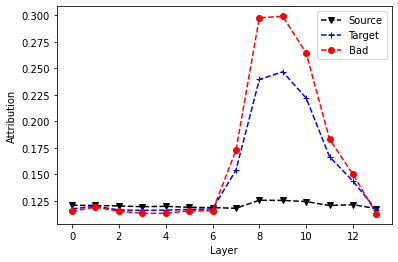}
\end{subfigure}%
\begin{subfigure}[t]{0.5\textwidth}
    \includegraphics[width=.9\textwidth]{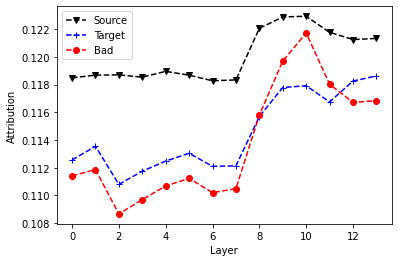}
\end{subfigure}
\caption{Average attribution at each hidden layer on the toy task (left) and MLQE-PE Et-En dataset (right). Attributions are computed with the information bottleneck attribution method \cite{schulz2020restricting}.}
\label{fig:attributions_across_layers}
\end{figure*}

\section{Experimental Settings}

\subsection{Evaluation}
\label{subsec:evaluation}

We start by introducing the evaluation metrics for assessing the performance of our approach. Given a test set with both sentence-level and word-level gold labels, we want to test to what extent the words with the highest attributions according to the QE model correspond to human annotation. Note that we cannot use the evaluation metrics traditionally employed for assessing the performance of word-level QE, such as F1 score and Matthews correlation coefficient \cite{specia-etal-2020-findings-wmt}, as they require binary predictions while feature attribution methods return continuous scores. Instead, we rely on metrics based on class probabilities instead \cite{atanasova2020diagnostic}. Since attribution methods proceed on instance-by-instance basis, and the scores produced for different instances are not necessarily comparable we compute the evaluation metrics for each instance separately and average the results across all instances in the test set.


\paragraph{AUC score} For each instance, we compute the area under the receiver operating characteristic curve (AUC score) to compare the continuous attribution scores $\vec{a}$ against binary gold labels $\vec{w}$. For a test set with $N$ instances:

\begin{equation}
AUC = \frac{1}{N}\sum_n{AUC_n(\vec{w}_n,\vec{a}_n^{\vec{x}^T})}
\end{equation}


\paragraph{Average Precision} AUC score can be overly optimistic for imbalanced data. Therefore, we also use Average Precision (AP).


\paragraph{Recall at Top-K} In addition, we report the Recall-at-Top-K metric commonly used in information retrieval. Applied to our setting, this metric computes the proportion of words with the highest attribution that correspond to translation errors against the total number of errors in the MT output. Thus, for a given instance (we omit the instance index $n$ here for simplicity):

\begin{equation}
\mathrm{Rec@TopK} = \frac{1}{k}\sum_{j\in \vec{e}_{1:k}}{\vec{w}_{j}}
\end{equation}

Where $\vec{e} = argsort(\vec{a}^{\vec{x}^T})$ is a sequence of indexes that sorts target words according to the attribution score from highest to lowest, and $k$ is the number of errors in the sentence. We then average the result across all instances in the test set.

\paragraph{Accuracy at Top-1} Finally, we report the proportion of sentences where the word with the highest attribution in the target corresponds to translation error. 

\begin{equation}
\mathrm{Acc@Top1} = \frac{1}{N}\sum{I[\vec{a}_{\vec{e}_1}=1]}
\end{equation}

We note that the above metrics are not defined for sentences where all words are labelled as errors or correct. We exclude such sentences from evaluation. 



\subsection{Sentence-level QE}

\begin{table}[t]
\begin{center}
\footnotesize
\begin{tabular}{l  c c c  } 
\toprule 
& {\bf Ro-En }& {\bf Et-En} &{\bf Ne-En} \\ 
\midrule
Pearson r & 0.84 & 0.66 & 0.66\\
Average DA & 68.9 & 55.2 & 36.6 \\
\midrule
Num. sentences total & 1,000 & 1,000 & 1,000 \\
Num. sentences (DA$<$70) & 438 & 640 & 935 \\
\midrule
Error rate (all data) & 0.21 & 0.28 & 0.65 \\
Error rate (DA$<$70) & 0.35 & 0.36 & 0.66 \\
\bottomrule
\end{tabular}
\end{center}
\caption{General statistics for MLQE-PE test sets: performance of sentence-level QE models (Pearson r), average DA score, total number of sentences in the test set, number of sentences with DA < 70, as well as error rate in the full test set and in the subset of selected sentences.}
\label{tab:statistics}
\end{table}

\renewcommand{\arraystretch}{1.1}\begin{table*}[t]\begin{center}\small\begin{tabular}{l  c c c c c c c c c c c c}
\toprule& \multicolumn{4}{c}{\bf Romanian-English} & \multicolumn{4}{c}{\bf Estonian-English} & \multicolumn{4}{c}{\bf Nepalese-English}\\
\cmidrule(r){2-5}\cmidrule(lr){6-9}\cmidrule(l){10-13}{\bf Method} & AUC & AP & A@1 & R@K & AUC & AP & A@1 & R@K & AUC & AP & A@1 & R@K \\
Gradients & 0.75 & 0.72 & \textbf{0.84} & 0.62 & \textbf{0.66} & \textbf{0.63} & \textbf{0.72} & \textbf{0.52} & 0.66 & 0.81 & \textbf{0.91} & 0.72\\
Info. Bottleneck & 0.65 & 0.62 & 0.71 & 0.50 & 0.58 & 0.55 & 0.56 & 0.46 & 0.64 & 0.78 & 0.80 & 0.71 \\
Attention & \textbf{0.79} & \textbf{0.73} & 0.80 & \textbf{0.63} & 0.65 & 0.57 & 0.52 & 0.49 & \textbf{0.69} & \textbf{0.82} & 0.88 & \textbf{0.74} \\
LIME & 0.54 & 0.48 & 0.40 & 0.39 & 0.56 & 0.56 & 0.65 & 0.46 & 0.52 & 0.75 & 0.76 & 0.68\\
\midrule
Random & 0.50 & 0.43 & 0.36 & 0.33 & 0.50 & 0.47 & 0.38 & 0.37 & 0.50 & 0.70 & 0.62 & 0.65\\
\midrule
Glassbox & 0.74 & 0.66 & 0.66 & 0.55 & 0.69 & 0.63 & 0.65 & 0.54 & 0.64 & 0.79 & 0.78 & 0.73\\
MicroTransQuest & 0.88 & 0.81 & 0.88 & 0.70 & 0.84 & 0.80 & 0.89 & 0.70 & 0.82 & 0.89 & 0.96 & 0.82\\
\bottomrule
\end{tabular}
\end{center}
\caption{AUC/AP scores, as well as accuracy at top-1 (A@1) and recall at top-K (R@K) for different rationale extraction methods on the test partition of MLQE-PE dataset. Best rationale extraction results are highlighted in bold.}
\label{tab:performance_mlqe_large}
\end{table*}


For sentence-level QE we rely on TransQuest \cite{ranasinghe2020transquest}, which was one of the top submissions to the WMT20 QE Shared Task \cite{specia-etal-2020-findings-wmt}. To facilitate the use of feature attribution methods described above we use our own implementation of the approach proposed by \cite{ranasinghe2020transquest,ranasinghe2020transquestwmt}. It achieves comparable results to the ones reported by the authors. Due to limited computational resources we use the XLM-R-base as underlying pre-trained Transformer model. We expect that using a more powerful sentence-level model would result in higher performances. 

\subsection{Data}

\begin{figure}[t]
\centering
\includegraphics[scale=0.55]{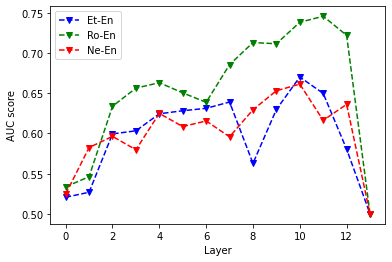}
\caption{AUC score at each hidden layer for integrated gradients method.}
\label{fig:score_per_layer}
\end{figure}

We use MLQE-PE (Multilingual Quality Estimation and Post-Editing) dataset described in \cite{fomicheva2020mlqe}.\footnote{\url{https://github.com/sheffieldnlp/mlqe-pe}} This dataset provides various types of manual MT evaluation for multiple language pairs. The MT outputs were assigned a sentence-level score inspired by the so called Direct Assessment annotation \cite{da_paper,guzman2019flores} on a continuous [0..100] scale capturing overall translation quality. 
In addition, the MT outputs were independently post-edited by professional translators. MT outputs and their corresponding post-edited versions were automatically aligned in order to derive word-level binary labels ("BAD" if the word was corrected, and "OK" otherwise), as well as the so called HTER score that corresponds to the average number of "BAD" labels in a sentence \cite{snover2006study}. We use these labels to evaluate the performance of different feature attribution approaches. We treat "BAD" labels as positive class and "OK" labels as negative class in our experiments.\footnote{The tokenization used internally by XLM-Roberta model is different from the tokenization used for producing word-level error labels. To map the attribution scores to the word labels we take their maximum value.} We do not evaluate attribution to source words. 



It is worth noting that word-level labels derived from post-editing do not capture error severity and do not always correspond to translation errors. However, due to the costs of collecting detailed error annotation for the amounts of data required to train SOTA models, this is a standard way of approximating error annotation in QE \cite{specia-etal-2020-findings-wmt}. \footnote{We have chosen this dataset as (i) it provides sufficient amount of word-level training data, which allows us to compare our approach to a SOTA supervised approach; and (ii) provides access to the neural MT models that were used to produce the translations, thus enabling a comparison to an unsupervised glass-box approach.} 

To circumvent the above limitation we leverage both types of sentence-level annotation (DA and HTER scores) in our experiments. We train sentence-level QE models with (i) DA scores and (ii) HTER scores. We evaluate both types of models using the word-labels derived from post-editing as described above. We then conduct evaluation as follows:
\begin{enumerate}
\item Evaluate explanations for DA-based models on the sentences with a sentence-level DA score lower than 70.\footnote{This threshold is selected based on the annotation guidelines described in \citet{fomicheva2020mlqe}, as the sentences assigned a score lower than 70 are guaranteed to have translation errors.}
\item Evaluate explanations for DA-based sentence-level models on the full subset of sentences that contain at least one word-level error.
\item Evaluate explanations for HTER-based sentence-level models on the full subset of sentences that contain at least one word-level error.
\end{enumerate}

Interestingly, despite the discrepancy between DA training objective and word labels derived from post-editing, explanations for DA-based models achieve better accuracy. We report the results for (1) in the main body of the paper, while (2) and (3) are reported in the Appendix.

We select three language pairs for our experiments: Et-En, Ro-En and Ne-En with the best performance at sentence level achieved at WMT2020 Shared Task. 
Table \ref{tab:statistics} shows various statistics for the respective test sets. These three language pairs present very different conditions for the task. Sentence-level model for Ro-En has much stronger performance in terms of Pearson correlation with human judgements. Ne-En has substantially lower translation quality where "BAD" words actually represent the majority class.

\subsection{Benchmarks}

\begin{figure}[t]
\begin{subfigure}[t]{0.5\columnwidth}
    \includegraphics[width=\linewidth]{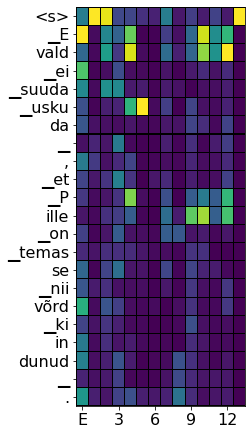}
\end{subfigure}%
\begin{subfigure}[t]{0.5\columnwidth}
    \includegraphics[width=\linewidth]{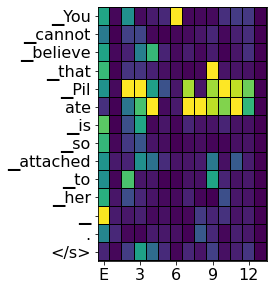}
\end{subfigure}
\caption{Example of Estonian-English translation with attributions to the source (left) and target (right) sentences computed using integrated gradients method for each hidden layer. The correct post-edited version of this translation is: \textit{\textbf{Evald} cannot believe that \textbf{Pille} is so attached to her.}}
\label{fig:example}
\end{figure}

\begin{figure*}[t]
\centering
\begin{subfigure}[t]{0.5\textwidth}
    \includegraphics[width=.9\textwidth]{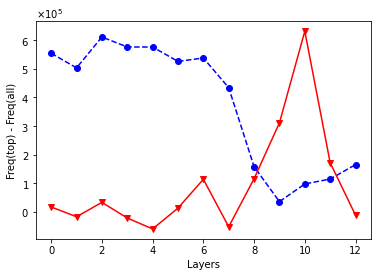}
\end{subfigure}%
\begin{subfigure}[t]{0.5\textwidth}
    \includegraphics[width=.9\textwidth]{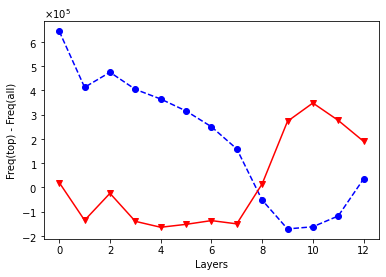}
\end{subfigure}
\caption{Frequency of the tokens with highest attribution in the neural MT training corpus. Y-axis shows the difference between average token frequency and frequency of the source (left) and target (right) tokens with the highest attribution scores in low-quality MT sentences (blue) and high-quality MT sentences (red). X-axis corresponds to the hidden layers.}
\label{fig:token_frequency}
\end{figure*}

We consider two benchmarks for word-level QE. On the one hand, we report the results for a strong supervised based on pre-trained representations from XLM-R but trained to predict word-level binary labels derived from post-editing. To report the metrics presented in \ref{subsec:evaluation}, we use probability of the positive class as attribution scores. On the other hand, we consider a fully unsupervised approach, which however, requires access to the neural MT model, that was used to generate the translations.

\paragraph{Black-box Supervised QE}
We use the word-level architecture available as part of the TransQuest toolkit \cite{ranasinghe2020transquest}\footnote{\url{https://tharindudr.github.io/TransQuest/architectures/word_level_architecture}}. Similarly to the sentence-level TransQuest model, it relies on XLM-Roberta-base pre-trained model fine-tuned for token classification task. We use XLM-Roberta-base to be consistent with the sentence-level settings. 

\paragraph{Glass-box Unsupervised QE}
Recently, \citet{fomicheva2020unsupervised} proposed to extract information from the neural MT system to predict translation quality in a fully unsupervised way. Following their work, we extract log-probabilities from the neural MT model and use them as attribution scores. The lower the log-probability corresponding to each word, the higher the chance that this word constitutes an error.

\subsection{Results}

Table \ref{tab:performance_mlqe_large} shows the performance of our approach with different attribution methods, as well as SOTA word-level QE methods for the MLQE-PE dataset. For the first three methods we compute the attributions to the hidden states at each layer on the dev set and report the results for this layer on the test set. First, the performance of our semi-supervised method with integrated gradients feature attribution approaches the performance of the supervised model for the Ro-En language pair, where the sentence-level QE model is the strongest (see Table \ref{tab:statistics})\footnote{The smallest gap with respect to the random baseline is observed for the Ne-En language pair. The overall quality of the translation for Ne-En is low. This setting is might be less suitable for the proposed error detection methods as most of the words in the data correspond to errors, as shown in Table \ref{tab:statistics}}. Second, our method performs comparably or better than the glass-box unsupervised method without requiring access to the neural MT model. Third, on average, LIME is substantially outperformed by the integrated gradients method. This agrees with our intuition that for the translation task where context plays a fundamental role, attribution to hidden states provides much better performance than direct perturbation of input words.


Figure \ref{fig:attributions_across_layers} shows attributions to tokens of different types across hidden layers. On the left, we show the results for a toy task, where we artificially introduced easy-to-detect errors in human translations and trained a QE model with near-perfect performance to predict whether a given sentence contain errors (see Appendix). On the right, we show the results for the the MLQE-PE Et-En test set. Similarly to the toy task, we observe that in the later layers the tokens corresponding to translation errors receive higher attribution scores. However, in the toy dataset, the source tokens have very low attributions. Here, in contrast, the model appears to be relying on the source as well as the target. This aligns very well with human evaluation where both source and target sentences need to be considered in order to correctly determine translation quality. 

Figure \ref{fig:score_per_layer} shows performance across layers for the integrated gradients method. As expected, the same layers that assign the highest attribution to the bad tokens (layers 9-11) are the ones that achieve the best performance. This finding is consistent across language pairs and attribution methods. Interestingly, this is also consistent with the findings reported in \citet{voita2019bottom}, where they show that models trained with MLM objective encode context information in intermediate layers partially discarding the information on the identity of the input tokens which is recovered at the latest layers. 

Figure \ref{fig:example} shows an example. Attributions are shown for sentencepiece tokens, which is the representation used internally by XLM-Roberta. Interestingly, both translation errors ("You" and "Pilate") and the corresponding words in the source ("Evald" and "Pille") receive higher attribution scores.


So far we have studied the behavior of the QE models on the sentences that contain errors. We now look at the pattern in the attributions scores for sentences which were assigned high quality by the model. We hypothesize that higher scores will be assigned to the words that are "easy" to translate. To test this, we select high-quality and low-quality sentences (sentences with predicted scores lower than 0.25 percentile and higher then 0.75 percentile, respectively). Figure \ref{fig:token_frequency} shows the average frequency with which the words occur in the neural MT training dataset. Red line corresponds to the words with the highest attribution for \textit{high-quality} MT sentences. Blue line corresponds to the words with the highest attribution for the \textit{low-quality} MT sentences. 
The first plot corresponds to the source tokens and the second plot corresponds to the target tokens. As shown in the plots, when the model predicts high quality the most frequent words receive the highest attribution as the information progresses through the network. By contrast when low quality is predicted by the sentence-level model, the least frequent words receive the highest attribution.


\section{Conclusion}
In this work we propose a new semi-supervised approach for word-level QE by exploring feature attribution methods. We show that for well performing models our results approach supervised performances. We introduce the QE task as a new benchmark for plausibility-based evaluation of rationale extraction methods. We hope this work will encourage further research on improving the efficiency of word-level QE models with lightly supervised methods. This work opens many directions for future research: from improving the achieved results by tuning linear weights to combine attributions to hidden states at different layers, to exploring different underlying architectures and sentence-level training objectives.

\section*{Acknowledgements}
This work was supported by funding from the Bergamot project (EU H2020 grant no. 825303).

\bibliography{custom}
\bibliographystyle{acl_natbib}

\appendix
\section{Toy dataset}

We devise a toy task to test feature attribution performance for word-level QE. We artificially introduce easy-to-detect errors in human translations and train a QE model with near-perfect performance to predict the presence/absence of such errors in a sentence. Specifically, we sample 10K/1K/1K sentence pairs from Es-En News-Commentary dataset (train/dev/test). Next, we artificially inject errors to half of the sentences at a rate of 0.1 using the following operations: insert, delete or replace random word, or swap two words selected at random. 

We fine-tune an XLM-R-base model for a sentence-level binary classification task where sentences that contain errors are considered as positive class, and sentences that do not contain errors are considered as negative class. The F1-score of this sentence-level classifier is 0.97. This is expected as the task is very easy.

\section{Performance of Rationale Extraction Methods on HTER Data}
\label{sec:hter}

Tables \ref{tab:performance_mlqe_da} and \ref{tab:performance_mlqe_hter} show the performance of the proposed methods on the full subset of sentences that contain at least one word-level error for sentence-level QE models trained with HTER and DA ground truth scores. Pearson correlation for both types of models is shown in Table \ref{tab:statistics_short}. Interestingly, even though for Ro-En and Et-En the performance of sentence-level models is near identical, extracted rationales are more accurate for the model trained with DA judgements.

\begin{table}[h!]
\begin{center}
\footnotesize
\begin{tabular}{l  c c c  } 
\toprule 
& {\bf Ro-En }& {\bf Et-En} &{\bf Ne-En} \\ 
\midrule
Pearson r (DA) & 0.84 & 0.66 & 0.66\\
Pearson r (HTER) & 0.82 & 0.62 & 0.51\\
\midrule
Num. sentences total & 1,000 & 1,000 & 1,000 \\
Num. sentences (DA$<$70) & 714 & 889 & 945 \\
\midrule
Error rate (all data) & 0.21 & 0.28 & 0.65 \\
Error rate (DA$<$70) & 0.28 & 0.31 & 0.65 \\
\bottomrule
\end{tabular}
\end{center}
\caption{Statistics for MLQE-PE test sets: performance of sentence-level QE models (Pearson r), total number of sentences with at least one translation error, and the error rate in the full test set and in the subset of selected sentences.}
\label{tab:statistics_short}
\end{table}

\renewcommand{\arraystretch}{1.1}\begin{table*}[t]\begin{center}\small\begin{tabular}{l  c c c c c c c c c c c c}
\toprule& \multicolumn{4}{c}{\bf Romanian-English} & \multicolumn{4}{c}{\bf Estonian-English} & \multicolumn{4}{c}{\bf Nepalese-English}\\
\cmidrule(r){2-5}\cmidrule(lr){6-9}\cmidrule(l){10-13}{\bf Method} & AUC & AP & A@1 & R@K & AUC & AP & A@1 & R@K & AUC & AP & A@1 & R@K \\
Gradients & 0.73 & \textbf{0.65} & \textbf{0.72} & \textbf{0.54} & \textbf{0.64} & \textbf{0.56} & \textbf{0.61} & \textbf{0.45} & 0.66 & \textbf{0.81} & \textbf{0.90} & 0.71\\
Info. Bottleneck & 0.59  & 0.49 & 0.50 & 0.36 & 0.54 & 0.47 & 0.42 & 0.37 & 0.62 & 0.76 & 0.78 & 0.69 \\
Attention & \textbf{0.76} & \textbf{0.65} & 0.67 & 0.53 & 0.63 & 0.51 & 0.45 & 0.41 & \textbf{0.69} & \textbf{0.81} & 0.87 & \textbf{0.73} \\
LIME & 0.51 & 0.39 & 0.29 & 0.29 & 0.55 & 0.49 & 0.54 & 0.39 & 0.52 & 0.73 & 0.72 & 0.66\\
\midrule
Random & 0.50 & 0.38 & 0.27 & 0.25 & 0.50 & 0.41 & 0.34 & 0.31 & 0.50 & 0.70 & 0.63 & 0.64\\
\midrule
Glassbox & 0.73 & 0.59 & 0.55 & 0.48 & 0.70 & 0.58 & 0.59 & 0.48 & 0.64 & 0.78 & 0.77 & 0.72\\
MicroTransQuest & 0.86 & 0.74 & 0.76 & 0.62 & 0.83 & 0.74 & 0.79 & 0.64 & 0.82 & 0.89 & 0.96 & 0.82\\
\bottomrule
\end{tabular}
\end{center}
\caption{AUC/AUPRC scores, as well as accuracy at top-1 (A@1) and recall at top-K (R@K) for different rationale extraction methods on the MLQE-PE test set on the subset of sentences that contain at least one error for the sentence-level QE models trained to predict DA judgements.}
\label{tab:performance_mlqe_da}
\end{table*}

\renewcommand{\arraystretch}{1.1}\begin{table*}[t]\begin{center}\small\begin{tabular}{l  c c c c c c c c c c c c}
\toprule& \multicolumn{4}{c}{\bf Romanian-English} & \multicolumn{4}{c}{\bf Estonian-English} & \multicolumn{4}{c}{\bf Nepalese-English}\\
\cmidrule(r){2-5}\cmidrule(lr){6-9}\cmidrule(l){10-13}{\bf Method} & AUC & AP & A@1 & R@K & AUC & AP & A@1 & R@K & AUC & AP & A@1 & R@K \\
Gradients & 0.69 & 0.59 & \textbf{0.61} & 0.48 & 0.66 & \textbf{0.59} & \textbf{0.66} & \textbf{0.49} & 0.64 & 0.77 & \textbf{0.82} & 0.70\\
Info. Bottleneck & 0.53 & 0.43 & 0.38 & 0.32 & 0.58 & 0.50 & 0.47 & 0.38 & 0.57 & 0.73 & 0.68 & 0.67 \\
Attention & \textbf{0.74} & \textbf{0.61} & 0.59 & \textbf{0.49} & \textbf{0.69} & \textbf{0.59} & 0.58 & 0.48 & \textbf{0.66} & \textbf{0.78} & \textbf{0.82} & \textbf{0.72} \\
LIME & 0.61 & 0.47 & 0.37 & 0.35 & 0.64 & 0.56 & 0.59 & 0.45 & 0.53 & 0.74 & 0.76 & 0.68\\
\midrule
Random & 0.50 & 0.38 & 0.27 & 0.25 & 0.50 & 0.41 & 0.33 & 0.32 & 0.50 & 0.70 & 0.63 & 0.64\\
\midrule
Glassbox & 0.73 & 0.59 & 0.55 & 0.48 & 0.70 & 0.58 & 0.59 & 0.48 & 0.64 & 0.78 & 0.77 & 0.72\\
MicroTransQuest & 0.86 & 0.74 & 0.76 & 0.62 & 0.83 & 0.74 & 0.79 & 0.64 & 0.82 & 0.89 & 0.96 & 0.82\\
\bottomrule
\end{tabular}
\end{center}
\caption{AUC/AUPRC scores, as well as accuracy at top-1 (A@1) and recall at top-K (R@K) for different rationale extraction methods on the MLQE-PE test set on the subset of sentences that contain at least one error for the sentence-level QE models trained to predict HTER.}
\label{tab:performance_mlqe_hter}
\end{table*}

\end{document}